\documentclass[11pt]{article}

\usepackage[margin=1in]{geometry}
\usepackage[T1]{fontenc}
\usepackage[utf8]{inputenc}
\usepackage{lmodern}
\usepackage{graphicx}
\usepackage{booktabs}
\usepackage{array}
\usepackage{multirow}
\usepackage{threeparttable}
\usepackage{float}
\usepackage{caption}
\usepackage{authblk}
\usepackage[numbers,sort&compress]{natbib}
\usepackage[colorlinks=true,linkcolor=blue,citecolor=blue,urlcolor=blue]{hyperref}

\title{Performance Evaluation of Open-Source Large Language Models for Assisting Pathology Report Writing in Japanese}

\author[1, 2]{Masataka Kawai\thanks{Corresponding author.}}
\author[3]{Singo Sakashita}
\author[3,4]{Shumpei Ishikawa}
\author[5]{Shogo Watanabe}
\author[5]{Anna Matsuoka}
\author[6]{Mikio Sakurai}
\author[1]{Yasuto Fujimoto}
\author[1]{Yoshiyuki Takahara}
\author[1]{Atsushi Ohara}
\author[1]{Hirohiko Miyake}
\author[1]{Genichiro Ishii}

\affil[1]{Department of Pathology and Clinical Laboratories, National Cancer Center Hospital East, Kashiwa, Japan}
\affil[1]{Department of Pathology, University of Yamanashi, Chuo, Japan}
\affil[3]{Division of Pathology, Exploratory Oncology Research \& Clinical Trial Center, National Cancer Center, Kashiwa, Japan}
\affil[4]{Department of Preventive Medicine, Graduate School of Medicine, The University of Tokyo, Tokyo, Japan}
\affil[5]{Department of Medical Oncology, National Cancer Center Hospital East, Kashiwa, Japan}
\affil[6]{Department of Thoracic Surgery, National Cancer Center Hospital East, Kashiwa, Japan}
\date{}

\begin{document}
\maketitle

\begin{abstract}
The performance of large language models (LLMs) for supporting pathology report writing in Japanese remains unexplored. We evaluated seven open-source LLMs from three perspectives: (A) generation and information extraction of pathology diagnosis text following predefined formats, (B) correction of typographical errors in Japanese pathology reports, and (C) subjective evaluation of model-generated explanatory text by pathologists and clinicians. Thinking models and medical-specialized models showed advantages in structured reporting tasks that required reasoning and in typo correction. In contrast, preferences for explanatory outputs varied substantially across raters. Although the utility of LLMs differed by task, our findings suggest that open-source LLMs can be useful for assisting Japanese pathology report writing in limited but clinically relevant scenarios.
\end{abstract}


\section{Introduction}
Since around 2022, large language models (LLMs) exemplified by ChatGPT have spread rapidly and are increasingly being considered for medical applications. Pathology reporting is one such domain, where LLMs can support report formatting, report correction, and explanatory writing. However, clinical deployment raises practical concerns regarding accuracy, fit to local workflow, reproducibility, and information governance.

Commercial cloud-based models such as ChatGPT, Gemini, and Claude offer strong performance, but using them with real pathology reports containing patient information is often difficult without dedicated contractual and governance arrangements. One practical alternative is to run open-source LLMs locally. Prior work has explored LLM-based information extraction and structured reporting in pathology, but evaluation methods remain heterogeneous and most reports focus on English-language material \citep{truhn_2023_extracting, rajaganapathy_2025_synoptic}. From the motivations above, we benchmarked open-source LLMs for assisting Japanese pathology report writing from three viewpoints: (A) formatting and extraction of breast pathology reports, (B) typo correction in real pathology reports, and (C) subjective evaluation of model-generated diagnostic explanations by pathologists and clinicians.

\section{Materials and Methods}
\subsection{Models and execution environment}
We downloaded models available through Hugging Face (https://huggingface.co) and served them through \texttt{llama.cpp} release b7640 using the \texttt{llama-server} OpenAI-compatible \texttt{/v1/chat/completions} API. We evaluated seven models available in January 2026: Gemma 3-27b-it (G3-27b) \citep{team_2025_gemma}, MedGemma-27b-text-it (MG-27b) \citep{sellergren_2025_medgemma}, SIP-jmed-llm-3-8x13b-AC-32k-instruct (SIP-jmed) \citep{nakamura_2025_dropupcycling}, Qwen3-Next-80B-A3B-Instruct (Q3N-In) and Qwen3-Next-80B-A3B-Thinking (Q3N-Th) \citep{yang_2025_qwen3}, and gpt-oss-20b and gpt-oss-120b \citep{openai_2025_gptoss120b}. Q4 or Q8 quantized variants were used. All experiments were run on a Mac Studio with an M2 Ultra and 196~GB memory.

Table~\ref{tab:modelinfo} summarizes model size, parameter count, quantization, and throughput measured by \texttt{llama-bench}. Prompt processing speed (\texttt{pp512}) and token generation speed (\texttt{tg128}) are reported as tokens per second. 

\begin{table}[H]
\centering
\scriptsize
\setlength{\tabcolsep}{4pt}
\begin{threeparttable}
\caption{Model information and \texttt{llama-bench} results.}
\label{tab:modelinfo}
\begin{tabular}{p{2.8cm}p{1.4cm}c c r r r r}
\toprule
Model & Quantization & Thinking & Med. knowledge & Size (GB) & Params (B) & \texttt{pp512} & \texttt{tg128} \\
\midrule
Gemma 3-27b-it & Q4\_0 & No & No & 16.04 & 27.01 & 384.66 & 29.90 \\
MedGemma-27b-text-it & Q4\_K\_XL & No & Yes & 15.66 & 27.01 & 337.32 & 27.31 \\
SIP-jmed-llm-3-8x13b-AC-32k-instruct & Q8\_0 & No & Yes & 72.40 & 73.16 & 421.82 & 25.97 \\
Qwen3-Next-80B-A3B-Instruct & Q8\_0 & No & No & 78.98 & 79.67 & 647.65 & 24.97 \\
Qwen3-Next-80B-A3B-Thinking & Q8\_0 & Yes & No & 78.98 & 79.67 & 650.06 & 24.94 \\
gpt-oss-20b & MXFP4 & Yes & No & 11.27 & 20.91 & 2293.59 & 120.27 \\
gpt-oss-120b & MXFP4 & Yes & No & 59.02 & 116.83 & 1170.35 & 80.03 \\
\bottomrule
\end{tabular}
\begin{tablenotes}
\item Med. knowledge indicates additional medical knowledge post-training. Throughput units are tokens/s.
\end{tablenotes}
\end{threeparttable}
\end{table}

\subsection{Benchmark A: formatted report generation and information extraction}
Using the 19th edition of the Japanese Breast Cancer Society reporting rules (Breast Kiyaku)\citep{jbs_2025_breast}, we evaluated support for breast surgical pathology reporting in four tasks.

\textbf{A1: JSON to institutional template.} Randomly generated JSON records were converted to a hospital-specific formatted report.

\textbf{A2: JSON to institutional template with pT determination and score calculation.} We removed pT, nuclear grade, and histological grade labels from the input JSON and required the LLM to infer them from invasion diameter and individual score components.

\textbf{A3: JSON to guideline template.} We provided one example of a guideline-format report and one conversion example, then asked each model to convert JSON to the format specified by Breast Kiyaku.

\textbf{A4: Institutional template to JSON.} We generated structured report text from JSON and asked the model to recover the original JSON items.

For A1 to A3, we compared model outputs against the reference text using character-level 3-gram F1 and Jaccard scores. For A2, we additionally computed accuracy for pT, nuclear grade (NG), histological grade (HG), and exact correctness across all three judgments. For A4, we computed the item-level JSON match ratio.

\subsection{Benchmark B: typo correction}
We randomly sampled 31 pathology reports of 100 to 400 Japanese characters from reports created at National Cancer Center Hospital East in 2024. After morphological segmentation with MeCab using an extended pathology-report dictionary, we injected synthetic typos with probability 0.1, including character deletion, incorrect insertion, transposition, and kanji conversion errors. Before prompting the LLMs, we applied preprocessing to reduce notation variability such as half-width and full-width space inconsistencies. A pathologist (M.K.) manually counted true positives (TP), false positives (FP), false negatives (FN), and large deletions (LD), where LD indicated phrase-level or sentence-level omission or distortion. When a model changed a target incorrectly, the error was counted as both FP and FN.

\subsection{Benchmark C: subjective evaluation of explanatory text}
From pathology reports diagnosed in 2024, M.K. selected 23 cases with extensive immunohistochemical workup and asked each LLM to generate an explanation suitable for a newby resident. For reports in which chronology was inverted because of additional reports, the sequence was reordered before combining the report with clinical information. Five board-certified pathologists and three clinicians with at least five years of clinical experience independently rated blinded outputs from all seven models on a 1 to 5 scale:

\begin{itemize}
\item 5: usable as an explanation without revision and without medical error,
\item 4: usable after minor wording revision and without major medical error,
\item 3: medically mostly acceptable but requiring revision of structure, explanation, or examples,
\item 2: containing medically important errors or requiring major revision,
\item 1: not functioning as an explanation.
\end{itemize}

Scores of 3 to 5 were regarded as useful. We summarized ratings by subgroup and calculated intraclass correlation coefficients ICC(2,1) and ICC(2,k).

\subsection{Hyperparameters, code availability, and ethics}
For A1 to A4 and B, all models except gpt-oss-20b and gpt-oss-120b were run with \texttt{temperature=0.1} and \texttt{top\_p=1.0}. For C, we used the parameters recommended for each model in the corresponding Hugging Face model card. Experimental code and prompts are available at \url{https://github.com/enigmanx20/PathLMbench_JP}. The study was approved by the National Cancer Center ethics committee (protocol 2025-098). All included cases had comprehensive consent. The data was anonymized, and handled in accordance with the Declaration of Helsinki.

\section{Results}
\subsection{Structured reporting and extraction}
In A1, all models achieved near-perfect text overlap, and both G3-27b and MG-27b reproduced the reference output exactly while remaining relatively fast (Table~\ref{tab:a1}). In A2, non-thinking models produced high string overlap but poor reasoning accuracy for pT and grade calculation, often close to chance. In contrast, Q3N-Th and both gpt-oss models achieved high or perfect reasoning accuracy, showing a clear advantage for models with deliberate reasoning behavior, consistent with prior reports on zero-shot reasoning in LLMs \citep{kojima_2022_large} (Table~\ref{tab:a2}).

In A3, all models except SIP-jmed followed the guideline template well, but errors were still observed in branching decisions such as adipose tissue invasion and dermal invasion. Q3N-Th achieved perfect agreement in 98 evaluable cases, with two timeouts. In A4, extraction performance was high for all models except SIP-jmed, and exact JSON recovery was obtained by G3-27b, MG-27b, and Q3N-In. Thinking models occasionally introduced small formatting discrepancies such as punctuation differences (Tables~\ref{tab:a3} and \ref{tab:a4}).

\begin{table}[H]
\centering
\scriptsize
\setlength{\tabcolsep}{4pt}
\caption{Benchmark A1: JSON to institutional template.}
\label{tab:a1}
\begin{tabular}{lrrrrrr}
\toprule
Model & Macro F1 & Macro Jaccard & Micro F1 & Micro Jaccard & Avg time (s) & Max time (s) \\
\midrule
Gemma 3-27b-it & \textbf{1.000} & \textbf{1.000} & \textbf{1.000} & \textbf{1.000} & 7.92 & 9.04 \\
MedGemma-27b-text-it & \textbf{1.000} & \textbf{1.000} & \textbf{1.000} & \textbf{1.000} & 8.51 & 9.76 \\
SIP-jmed-llm-3-8x13b-AC-32k-instruct & 0.989 & 0.980 & 0.987 & 0.974 & 8.20 & 16.28 \\
Qwen3-Next-80B-A3B-Instruct & 0.999 & 0.998 & 0.999 & 0.998 & 11.13 & 11.98 \\
Qwen3-Next-80B-A3B-Thinking & \textbf{1.000} & \textbf{1.000} & \textbf{1.000} & \textbf{1.000} & 119.10 & 170.81 \\
gpt-oss-20b & 0.997 & 0.994 & 0.997 & 0.994 & 10.94 & 24.51 \\
gpt-oss-120b & 0.996 & 0.992 & 0.996 & 0.992 & 16.11 & 36.32 \\
\bottomrule
\end{tabular}
\end{table}

\begin{table}[H]
\centering
\scriptsize
\setlength{\tabcolsep}{4pt}
\caption{Benchmark A2: JSON to institutional template with pT determination and score calculation.}
\label{tab:a2}
\begin{tabular}{lrrrrrrrr}
\toprule
Model & Macro F1 & Macro Jaccard & Micro F1 & Micro Jaccard & pT & NG & HG & All correct \\
\midrule
Gemma 3-27b-it & 0.987 & 0.974 & 0.987 & 0.974 & 0.18 & 0.66 & 0.58 & 0.47 \\
MedGemma-27b-text-it & 0.983 & 0.966 & 0.983 & 0.966 & 0.06 & 0.41 & 0.51 & 0.36 \\
SIP-jmed-llm-3-8x13b-AC-32k-instruct & 0.428 & 0.277 & 0.430 & 0.274 & 0.00 & 0.00 & 0.00 & 0.01 \\
Qwen3-Next-80B-A3B-Instruct & 0.987 & 0.975 & 0.987 & 0.975 & 0.18 & 0.66 & 0.59 & 0.37 \\
Qwen3-Next-80B-A3B-Thinking & \textbf{1.000} & \textbf{1.000} & \textbf{1.000} & \textbf{1.000} & \textbf{1.00} & \textbf{1.00} & \textbf{1.00} & \textbf{1.00} \\
gpt-oss-20b & 0.991 & 0.983 & 0.991 & 0.982 & 0.96 & 0.97 & 0.98 & 0.99 \\
gpt-oss-120b & 0.995 & 0.989 & 0.994 & 0.989 & \textbf{1.00} & \textbf{1.00} & \textbf{1.00} & \textbf{1.00}  \\
\bottomrule
\end{tabular}
\end{table}

\begin{table}[H]
\centering
\scriptsize
\setlength{\tabcolsep}{4pt}
\begin{threeparttable}
\caption{Benchmark A3: JSON to Kiyaku template.}
\label{tab:a3}
\begin{tabular}{lrrrrrr}
\toprule
Model & Macro F1 & Macro Jaccard & Micro F1 & Micro Jaccard & Avg time (s) & Max time (s) \\
\midrule
Gemma 3-27b-it & 0.992 & 0.985 & 0.992 & 0.985 & 4.44 & 5.90 \\
MedGemma-27b-text-it & 0.993 & 0.986 & 0.993 & 0.986 & 4.65 & 7.07 \\
SIP-jmed-llm-3-8x13b-AC-32k-instruct & 0.418 & 0.301 & 0.390 & 0.242 & 9.06 & 10.34 \\
Qwen3-Next-80B-A3B-Instruct & 0.996 & 0.993 & 0.996 & 0.993 & 6.41 & 6.79 \\
Qwen3-Next-80B-A3B-Thinking\tnote{*} & \textbf{1.000} & \textbf{1.000} & \textbf{1.000} & \textbf{1.000} & 192.14 & 300.00 \\
gpt-oss-20b & 0.971 & 0.944 & 0.971 & 0.943 & 13.62 & 28.95 \\
gpt-oss-120b & 0.990 & 0.981 & 0.990 & 0.981 & 11.18 & 24.93 \\
\bottomrule
\end{tabular}
\begin{tablenotes}
\item[*] Results are based on 98 completed cases because of 2 timeouts.
\end{tablenotes}
\end{threeparttable}
\end{table}

\begin{table}[H]
\centering
\scriptsize
\setlength{\tabcolsep}{4pt}
\caption{Benchmark A4: extraction of JSON items from a formatted report.}
\label{tab:a4}
\begin{tabular}{lrrr}
\toprule
Model & Match ratio & Avg time (s) & Max time (s) \\
\midrule
Gemma 3-27b-it & \textbf{1.000} & 9.43 & 10.85 \\
MedGemma-27b-text-it & \textbf{1.000} & 10.27 & 12.19 \\
SIP-jmed-llm-3-8x13b-AC-32k-instruct & 0.532 & 9.64 & 11.42 \\
Qwen3-Next-80B-A3B-Instruct & \textbf{1.000} & 12.42 & 12.79 \\
Qwen3-Next-80B-A3B-Thinking & 0.955 & 178.64 & 456.17 \\
gpt-oss-20b & 0.984 & 11.27 & 18.73 \\
gpt-oss-120b & 0.986 & 15.41 & 33.65 \\
\bottomrule
\end{tabular}
\end{table}

\subsection{Typo correction}
Table~\ref{tab:typo} summarizes typo correction performance. Q3N-In achieved the best overall balance with a macro F1 of 0.697 and micro F1 of 0.717. SIP-jmed occasionally corrected difficult pathology-specific errors that other models missed, but it also produced the highest number of large deletions. MG-27b and Q3N-Th also showed competitive typo correction performance, whereas gpt-oss-20b performed worst overall in this task.

\begin{table}[H]
\centering
\scriptsize
\setlength{\tabcolsep}{4pt}
\caption{Benchmark B: typo correction in 31 pathology reports.P: presicion, R: Recall.}
\label{tab:typo}
\begin{tabular}{lrrrrrrr}
\toprule
Model & Macro P & Macro R & Macro F1 & Micro P & Micro R & Micro F1 & LD \\
\midrule
Gemma 3-27b-it & 0.763 & 0.557 & 0.619 & 0.759 & 0.538 & 0.630 & 1 \\
MedGemma-27b-text-it & 0.773 & 0.574 & 0.629 & 0.827 & 0.578 & 0.680 & 0 \\
SIP-jmed-llm-3-8x13b-AC-32k-instruct & 0.723 & 0.686 & 0.640 & 0.713 & 0.713 & 0.713 & 4 \\
Qwen3-Next-80B-A3B-Instruct & 0.738 & 0.673 & \textbf{0.697} & 0.743 & 0.692 & \textbf{0.717} & 0 \\
Qwen3-Next-80B-A3B-Thinking & 0.633 & 0.684 & 0.651 & 0.672 & 0.735 & 0.702 & 0 \\
gpt-oss-20b & 0.544 & 0.501 & 0.517 & 0.559 & 0.491 & 0.523 & 1 \\
gpt-oss-120b & 0.651 & 0.638 & 0.634 & 0.651 & 0.611 & 0.630 & 1 \\
\bottomrule
\end{tabular}
\end{table}

\subsection{Subjective evaluation of explanatory text}
\begin{figure}[H]
\centering
\includegraphics[width=\textwidth]{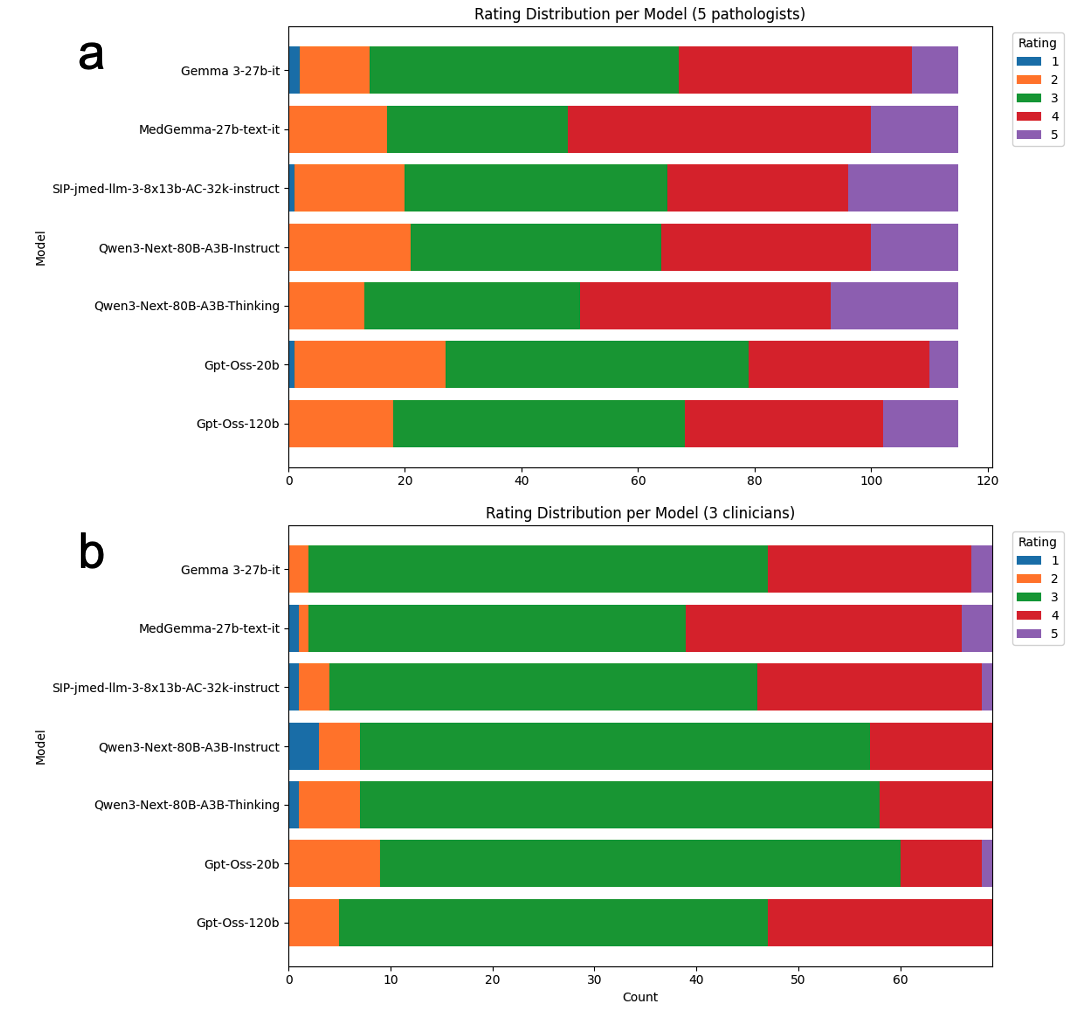}
\caption{Distribution of explanatory-text ratings by model among five pathologists (a) and three clinicians (b).}
\label{fig:ratings}
\end{figure}

Among pathologists and clinicians, roughly one quarter to one third of all ratings were 4 or 5, indicating outputs that were potentially useful with little or no revision (Figure~\ref{fig:ratings}). MG-27b also received many scores of 4 and 5, while Q3N-Th was rated more favorably by pathologists than by clinicians.

\begin{figure}[H]
\centering
\includegraphics[width=\textwidth]{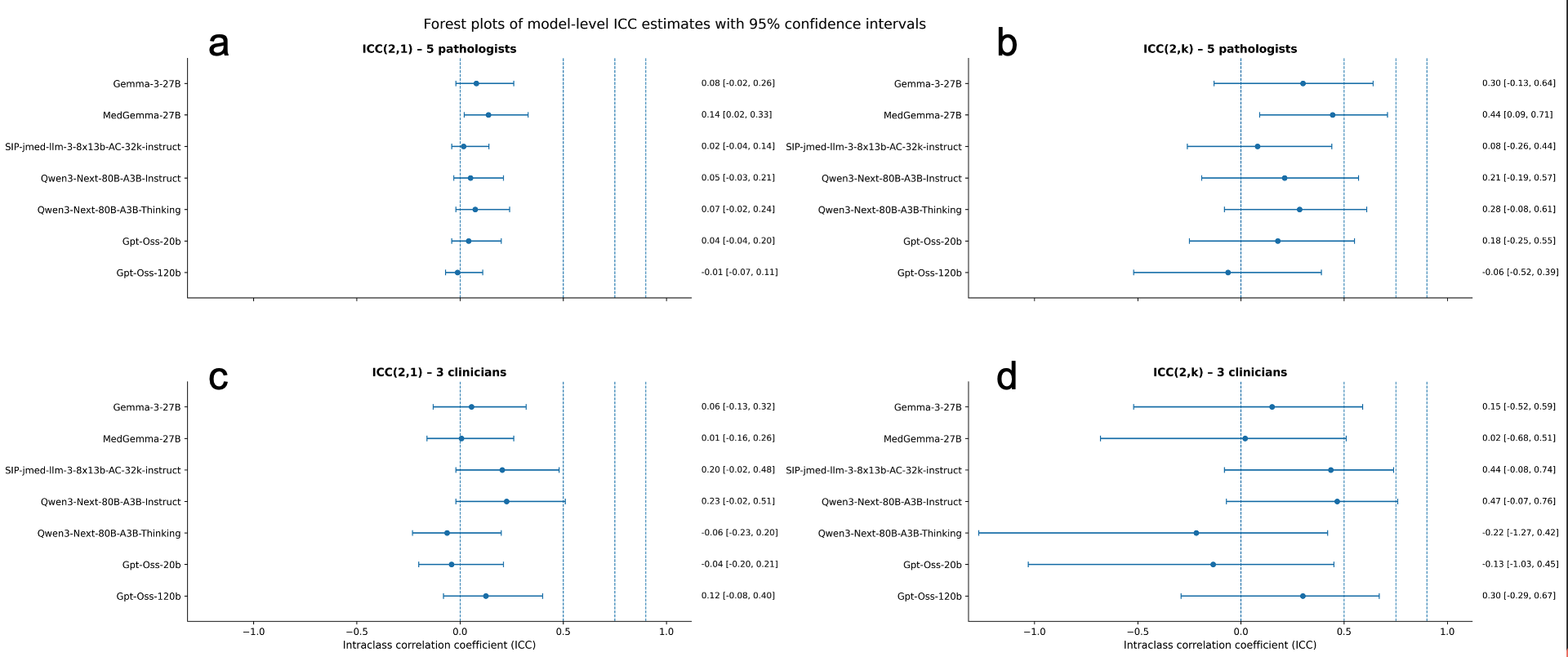}
\caption{Model-level inter-rater reliability for explanatory-text evaluation. Panels show ICC(2,1) and ICC(2, k) of pathologists (a and b). Those of clinicians (c and d). 95\% confidence intervals are depicted with bars.}
\label{fig:icc}
\end{figure}

Inter-rater agreement at the single-rater level was generally low (Figure~\ref{fig:icc}). For pathologists, the highest ICC(2,1) was 0.138 for MG-27b; for clinicians, the highest ICC(2,1) was 0.226 for Q3N-In. Reliability improved when using mean scores across raters: among pathologists, the highest ICC(2,k) was 0.444 for MG-27b; among clinicians, the highest ICC(2,k) was 0.466 for Q3N-In (Figure~\ref{fig:icc}). These findings indicate large variability in human preference for model-generated explanations.

\section{Discussion}
This study benchmarked seven open-source LLMs for assisting Japanese pathology report writing across formatting, extraction, typo correction, and explanatory writing tasks. No single model dominated all benchmarks. Instead, utility was task and model dependent.

For deterministic transformations such as JSON to template conversion, conventional programs remain faster and more reliable than LLMs, especially when the mapping rules are already known. By contrast, several use cases appear promising for open-source LLMs: conversion from an example-driven reporting style to a new format, typo correction, and generation of case explanations that help bridge the gap between pathology reports and non-pathologist readers. In particular, medical specialization appeared beneficial in typo correction and explanatory writing, as illustrated by MG-27b and SIP-jmed.

The subjective evaluation task also highlighted a major challenge: ratings varied widely across experts. This likely reflects both medical judgment and differences in stylistic preference, which is often the general observations about human preference variability in LLM evaluation \citep{petrova_2026_unpacking}. Future systems for pathology reporting support may therefore need to be personalized or customized to institutional and individual reporting preferences rather than optimized against a single universal preference target.

Compared with closed commercial models, open-source models offer advantages in privacy control, local deployment, and reproducibility, although they still require substantial hardware resources. At the time of evaluation in January 2026, practically useful off-the-shelf models generally had at least 10 billion parameters. Ongoing improvements in mixture-of-experts architectures, quantization, and reasoning-oriented training may reduce these hardware requirements over time.

Our benchmark also sits in the context of prior pathology-focused LLM evaluation studies, which have examined information extraction, structured data capture, and patient-facing report generation \citep{grothey_2025_comprehensive,proctor_2025_bridging}. More broadly, available pathology-specific benchmark resources remain limited and are mostly English-based \citep{he_2020_pathvqa}. This study has several limitations. First, we did not directly compare open-source models with commercial closed models such as GPT5.2. Second, reproducibility may be affected by prompt design and stochastic nature of sampling. Third, hyperparameter and prompt optimization were limited. Fourth, some models were evaluated in quantized form, which may have affected performance. Despite these limitations, the benchmark identifies clinically relevant task categories in which open-source LLMs may already be useful for Japanese pathology report support.

\section{Conclusion}
Open-source LLMs showed limited but meaningful utility for assisting pathology report writing in Japanese. Thinking models were especially strong when explicit reasoning was required, while medical-specialized models appeared advantageous for typo correction and explanatory writing. Because performance differed markedly by task and human preferences varied across evaluators, practical deployment will likely require task selection, local validation, and eventual personalization.

\section*{Acknowledgments}
ChatGPT, Gpt-Oss-120b, and Qwen3.5-27B were used during code development and manuscript writingg. The authors reviewed and approved the final manuscript and take full responsibility for its content.

\bibliographystyle{unsrtnat}
\bibliography{references}

\end{document}